\documentclass[letterpaper, 10 pt, conference]{ieeeconf}  

\IEEEoverridecommandlockouts                              

\usepackage{graphicx}
\usepackage{graphics} 
\usepackage{amssymb}  
\usepackage{amsmath}
\usepackage{url}
\usepackage{booktabs}
\usepackage{multirow}    
\usepackage{makecell}    
\usepackage{adjustbox}   
\usepackage{pifont}      
\usepackage[justification=raggedright,singlelinecheck=false]{caption}
\usepackage{booktabs} 
\usepackage{makecell} 

\usepackage{xcolor}
\newcommand{\tblscale}{0.86} 

\title{\LARGE \bf
ScanBot: A Benchmark for Precision Robotic Surface Scanning with Industrial Laser Profilers
}

\author{Zhiling Chen$^{1}$, Yang Zhang$^{1}$, Fardin Jalil Piran$^{1}$, Qianyu Zhou$^{1}$, Jiong Tang$^{1}$ and Farhad Imani$^{1}$
\thanks{$^{1}$Zhiling  are with the University of Connecticut
        {\tt\small {zhiling.chen, yang.3.zhang, fardin.piran, qianyu.zhou, jiong.tang, farhad.imani}@uconn.edu}}%
}

\author{Zhiling Chen$^{1}$, Yang Zhang$^{1}$, Fardin Jalil Piran$^{1}$, Qianyu Zhou$^{1}$, Jiong Tang$^{1}$, and Farhad Imani$^{1}$
\thanks{$^{1}$Zhiling Chen, Yang Zhang, Fardin Jalil Piran, Qianyu Zhou, Jiong Tang, and Farhad Imani are with the University of Connecticut
{\tt\small \{zhiling.chen, yang.3.zhang, fardin.piran, qianyu.zhou, jiong.tang, farhad.imani\}@uconn.edu}}%
}

\begin{document}

\maketitle
\thispagestyle{empty}
\pagestyle{empty}

\begin{figure*}[]
  \centering
  \includegraphics[width=\textwidth]{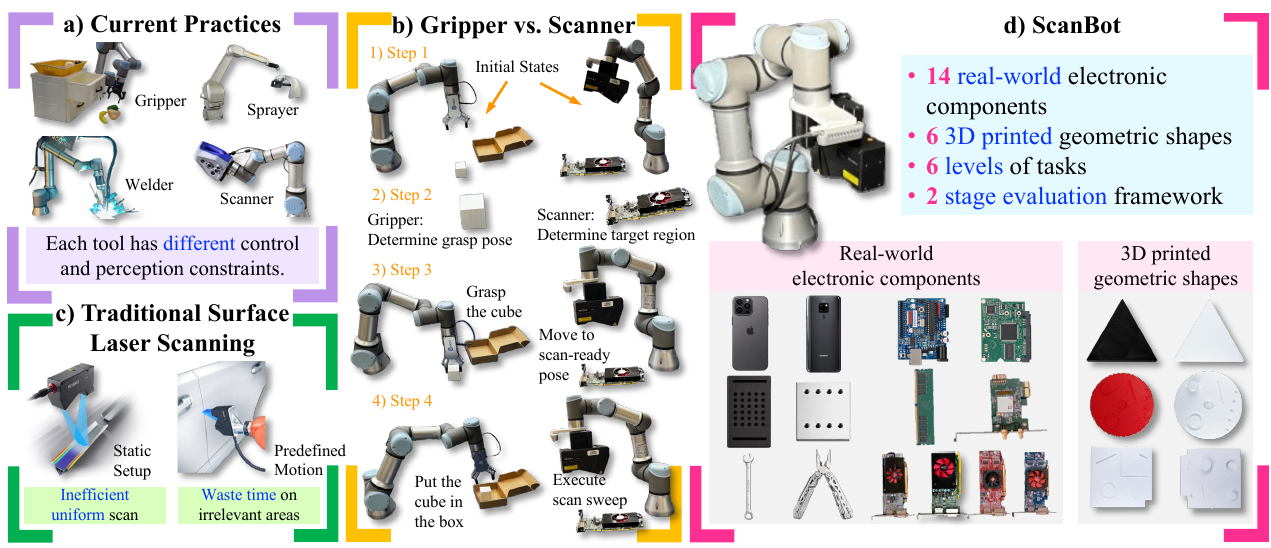} 
  \caption{Overview and motivation for the ScanBot benchmark. (a) Embodied AI should generalize across tools with distinct control and perception demands. (b) Unlike discrete gripper interactions, scanning requires precise region localization and smooth, continuous motion. (c) Conventional laser scanning uses fixed, task-agnostic paths, wasting time on irrelevant areas. (d) ScanBot includes 14 real components and 6 3D-printed shapes, supporting 6 task types and a 2 stage evaluation framework for instruction-conditioned surface scanning.}
  \label{introduction}
\end{figure*}

\begin{abstract}

We introduce ScanBot, a benchmark for instruction-conditioned, high-precision surface scanning with robot-mounted industrial laser profilers. Unlike existing robot learning datasets that emphasize coarse behaviors such as grasping, navigation, or dialogue, ScanBot targets sensing-centric tasks where sub-millimeter motion continuity, strict stand-off control, and stable scanner settings are essential for acquiring usable geometry. The dataset contains scanning trajectories over twenty objects, including electronic components and structured 3D-printed parts, and spans six task types that range from broad inspection to fine-grained detail scanning and geometry-critical operations, including metrology and registration. Each episode is specified by natural language instructions and paired with synchronized first-person RGB-D, third-person video, laser height profiles, robot joint and pose traces, and scanner-parameter logs. These requirements expose a gap: despite recent progress, learning-based models often fail to produce stable and feasible scan motions under fine-grained instructions and real laser-profiling constraints. To reflect how industrial scanning is actually done, we evaluate methods through a two-stage pipeline. Stage I asks the model to "set up the sensor" by recommending scanner parameters, while Stage II asks it to "move like a scanner" by producing smooth, feasible trajectories that maintain stand-off and cover the intended region under precision demands.

\end{abstract}

\section{INTRODUCTION}
\label{introduction text}
Embodied-AI systems have recently achieved strong results on tasks such as navigation, pick-and-place, and natural-language manipulation~\cite{kim2024openvla}. However, many real deployments revolve around tool-mediated sensing and actuation, where motion tolerances and parameter stability are far stricter than those faced by a parallel gripper. A paint sprayer must maintain a $\pm 2$\,mm standoff and $\pm 50$\,mm\,s$^{-1}$ sweep-speed to avoid orange-peel artifacts; a TIG torch must track a bead within $\pm 0.2$\,mm; and a laser line-scanner, our focus, must limit trajectory jitter to below $0.1$\,mm, matching the sensor’s depth resolution, as illustrated in Fig.~\ref{introduction}(a). Despite the centrality of such skills, there is no public benchmark that asks whether instruction-conditioned embodied models can execute sub-millimeter, time-parameterized trajectories while keeping sensing parameters within admissible bounds.


The impact of this gap is most pronounced in non-destructive evaluation of large, high-value components such as turbine blades and battery housings. Exhaustive full-surface scans scale quadratically with part size and squander bandwidth on regions that rarely fail. For example, scanning a $1$\,m $\times$ $0.5$\,m turbine blade at $0.1$\,mm resolution entails $\approx 50$ million points and more than $45$ minutes with a 1\,kHz sensor, which is untenable for in-line quality control. Practitioners therefore seek target-aware scanning, e.g., ``inspect the leading-edge weld'', where the robot grounds the phrase to a precise mesh region and adapts standoff, speed, and exposure on the fly. Mounting a scanner inside a gripper amplifies vibration and occlusion; rigid flange mounting mitigates those issues but converts a brief grasp into thousands of continuous waypoints whose geometry, timing, and illumination jointly determine data quality~\cite{al2021integrating}.


Fig.~\ref{introduction}(b) contrasts gripper manipulation and surface scanning in four steps. (1) Perception: both recognize the object, but a gripper reasons about a graspable solid, whereas a scanner reasons about a measurable surface. (2) Goal grounding: the gripper selects a grasp pose, while the scanner must ground the instruction to a specific region and choose admissible sensing parameters under visibility/stand-off constraints. (3) Approach: the gripper executes a short reach; the scanner must reach a scan-ready pose that meets stand-off and visibility requirements to avoid occlusion and measurement artifacts. (4) Execution and evaluation: the gripper finishes with a discrete pick/place, while scanning requires a smooth, time-continuous sweep with stable velocity and orientation; success is evaluated by coverage, geometric alignment, and motion stability (e.g., jitter/jerk/curvature spikes) rather than a binary completion signal.


Existing embodied datasets and models are poorly matched to these demands. Prior corpora either assume a static scanner observing a moving part~\cite{petrie2018introduction}, restrict sensing to sparse point clouds without joint-state supervision, or confine evaluation to success–fail task completion. As a result, widely used vision-language-action (VLA) models that excel at grasping or navigation offer little evidence of competence in precision scanning. Their abstractions reason over bounding boxes and discrete actions or sparse waypoints rather than the millimeter-sensitive, time-continuous pose curves with coupled tool-state constraints that are required for artifact-free scans.
This mismatch is mirrored in industrial practice. As depicted in Fig.~\ref{introduction}(c), laser scanning is typically executed either with a static sensor that views objects on a conveyor or with a robot following hard-coded, blanket-coverage trajectories. These pipelines can suffice for simple, uniform parts, but they become inefficient and brittle on realistic assemblies. Large components render exhaustive coverage prohibitively slow; defects cluster at connectors, weld seams, and edges; and many defects (e.g., micro-cracks, delaminations) are only detectable under tuned conditions, slower scan speeds, higher exposure, or tighter standoff, that cannot be applied uniformly without crippling throughput.


Motivated by these realities, we introduce ScanBot, an instruction-conditioned multimodal dataset designed explicitly for high-precision robotic surface scanning, as shown in Fig.~\ref{introduction}(d). ScanBot spans twenty objects, combining real electronic components with analytically structured 3D-printed shapes, each annotated with multiple task instructions and paired with high-resolution scanning trajectories. We define six representative scanning tasks for industrial quality inspection, including four appearance inspection tasks and two dimensional and geometric metrology tasks, spanning broad surface coverage to fine-grained local inspection and comparative alignment. Each example includes synchronized first-person RGB-D imagery, third-person overview video, laser height profiles, robot pose and joint states, and full sensor-parameter traces, providing supervision across perception, planning, and control. The dataset is organized around instruction grounding to geometry, parameter adherence during execution, and trajectory stability under realistic precision constraints, enabling reproducible comparisons.

To evaluate model performance on the ScanBot dataset, we introduce a benchmark that targets the full stack of precision scanning, spanning a two-stage evaluation protocol: Stage I for scanner-parameter selection, and Stage II for generating stable, feasible, and target-covering scan motions without extra moves. In contrast to manipulation-centric settings, surface scanning is highly sensitive to motion smoothness and sensing geometry: small deviations in stand-off distance, orientation, or path continuity can trigger measurement dropouts and rapidly degrade reconstruction quality. We therefore evaluate a diverse set of baselines across both stages, including an industrial fixed-template scanning routine, learning-based policies, and state-of-the-art VLA models, together with parameter-selection baselines ranging from rule-based heuristics and ablations to zero-shot multimodal models. Collectively, these experiments probe whether today’s models can move beyond coarse waypoint following and deliver the precise spatial grounding and stable control required by realistic industrial scanning scenarios.

\section{Related Works}

\subsection{Robot Learning Datasets}

The development of general-purpose robotic policies relies heavily on large-scale, diverse datasets~\cite{brohan2022rt, kumar2023robohive, fang2023rh20t}. Inspired by advances in vision and language pretraining~\cite{zhai2023sigmoidlosslanguageimage, tschannen2025siglip2multilingualvisionlanguage,oquab2023dinov2}, recent efforts emphasize collecting heterogeneous data across tasks, objects, and embodiments. Datasets like Open X-Embodiment~\cite{o2024open} and RoboMIND~\cite{wu2024robomind} exemplify this trend, aggregating diverse trajectories across robot types to enable broader generalization. However, most existing datasets remain focused on manipulation tasks using grippers or dexterous hands, overlooking sensor-centric, perception-driven activities. To fill this gap, we present a dataset centered on active perception using a robot-mounted 3D laser scanner. Unlike manipulation, sensor-guided tasks emphasize precise pose control to optimize coverage and measurement quality. This shifts the embodiment-task interface from object interaction to coverage-aware motion planning. Our dataset introduces robot-guided scanning trajectories, pose-aligned coverage metrics, and instruction-driven scan directives, complementing existing datasets while broadening robot learning from what to do to what to sense.

\subsection{Laser Scanner}

Laser scanners are widely used in industrial inspection to capture high-resolution surface measurements for defect detection, tolerance verification, and reverse engineering. While 2D line-scan systems are fast and common in inline settings, 3D laser profilers aggregate dense scan lines into full surfaces for shape analysis and quality assurance, motivating robot-mounted scanning for automation. Traditional robot scanning relies on manually programmed trajectories that are hard to adapt to new geometries. Recent work studies automated viewpoint planning and coverage path planning under sensor constraints, including region segmentation and global optimization~\cite{chen2025pso}, as well as learning-based approaches such as reinforcement learning for adaptive viewpoints~\cite{roos2025reinforcement}. However, existing methods are often domain-specific and evaluated under inconsistent protocols, and standardized benchmarks remain limited. ScanBot addresses this gap with an instruction-conditioned, multimodal dataset for high-resolution surface scanning, where the scanner interprets language, grounds target regions, selects parameters, and executes smooth trajectories.


\section{The ScanBot Benchmark}

Our goal is to develop a benchmark that enables research on tool-specific embodied perception beyond traditional manipulation. Unlike conventional embodied datasets that focus on discrete object interactions, ScanBot emphasizes high-precision, tool-mediated surface scanning under real-world sensing constraints. The benchmark is structured to promote generalization across diverse scan tasks, object geometries, and inspection objectives, while enabling instruction-conditioned policy learning where robots must interpret high-level language commands to generate feasible and stable scanning motions. In this section, we describe the design of ScanBot, including our data collection setup, task structure, and dataset composition.

\begin{figure}[]
  \centering
  \includegraphics[width=0.9\linewidth]{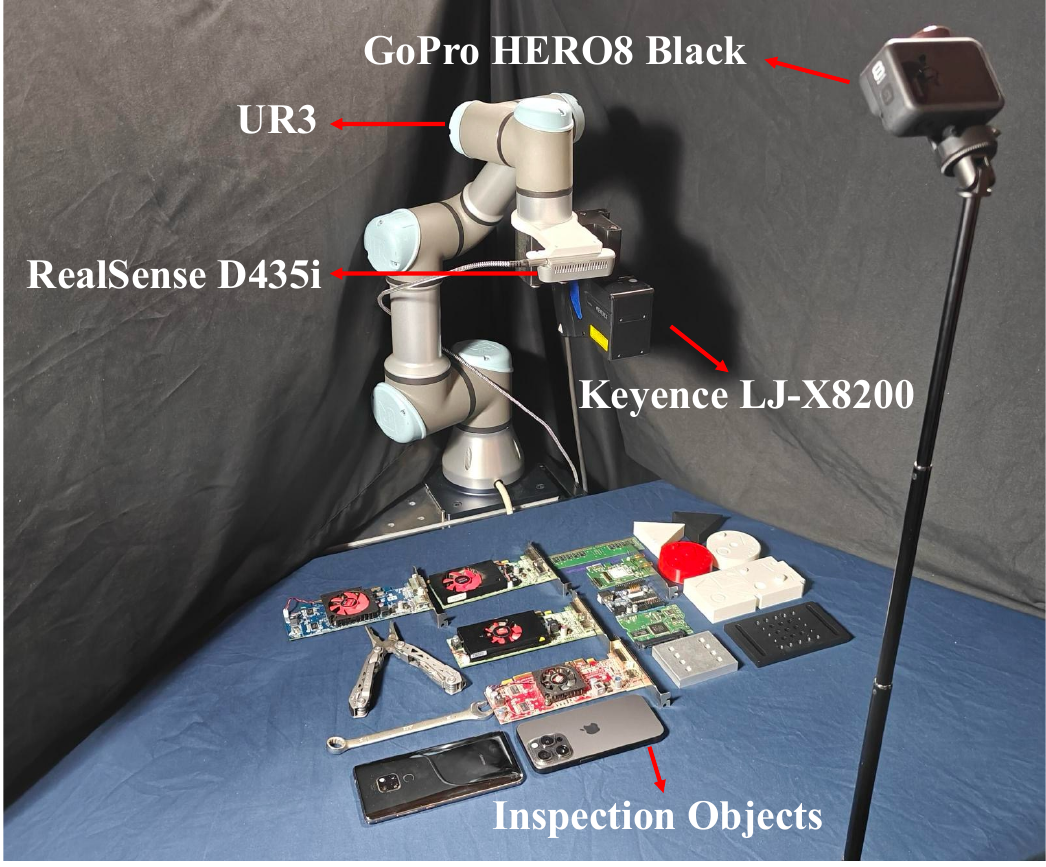}
  \caption{Hardware setup of the ScanBot system.}
  \label{setup}
\end{figure}

\subsection{System Overview and Task Definition}
\label{sec:problem_setup}

We consider the problem of instruction-conditioned robotic surface scanning, where a robot-mounted laser scanner configures its sensing parameters and executes a scan motion based on visual observations and a natural-language instruction.
Each scanning instance is defined by an observation--instruction pair $(s, w) \in \mathcal{S} \times \mathcal{W}$, where $s \in \mathcal{S}$ denotes first-person RGB or RGB-D observations captured from a wrist-mounted camera, and $w \in \mathcal{W}$ denotes a task command specifying the inspection intent. Different instructions induce different sensing regimes and scan extents.

The objective of ScanBot is to predict a structured scanning action $a \in \mathcal{A}$ conditioned on $(s, w)$, with the action space defined as $\mathcal{A} = \Theta \times \mathcal{T}$, where $\Theta$ denotes scanner configuration parameters and $\mathcal{T}$ denotes scanning motions. Although industrial laser scanners expose a large number of configurable options, many are strongly coupled or task-specific. To keep the benchmark tractable while preserving the dominant factors that affect scan quality, we restrict the configuration space to a small set of high-impact parameters supported by the Keyence LJ-X8000 laser profiling system.
Specifically, the scanner configuration is represented as a five-dimensional vector $\boldsymbol{\theta} = (f_s, r_x, r_c, t_e, r_l)$, where $f_s$ is the sampling frequency, $r_x$ is the lateral measurement range, $r_c$ is the CMOS dynamic range, $t_e$ is the exposure time, and $r_l$ is the control light intensity. The lateral range $r_x$ directly determines the effective width of each laser profile, with narrower ranges favoring fine-grained inspection and wider ranges enabling faster surface coverage. All parameters are selected prior to motion execution and remain fixed during each scan.
After configuring the scanner, the robot executes a scanning motion $\tau \in \mathcal{T}$ that implicitly determines the measured surface region. 
We adopt a simplified motion abstraction in which each scan is represented as a task-aligned straight-line sweep specified by a nominal tool pose $\mathbf{p}_{\mathrm{ref}} \in SE(3)$, a unit scan direction $\hat{\mathbf{d}}$, and a sweep extent $\ell$. The executed trajectory is a time-parameterized pose curve that follows this sweep while maintaining the required stand-off and tool alignment. 

Overall, ScanBot defines instruction-conditioned surface scanning as a conditional mapping $\mu : \mathcal{S} \times \mathcal{W} \rightarrow \mathcal{A}$ from observations and task instructions to scanner configurations and scan motions. We restrict attention to planar or locally planar surfaces and consider single-pass feedforward execution without online adaptation, enabling controlled and reproducible evaluation of sensing-driven embodied scanning.



\begin{figure*}[]
  \centering
  \includegraphics[width=\linewidth]{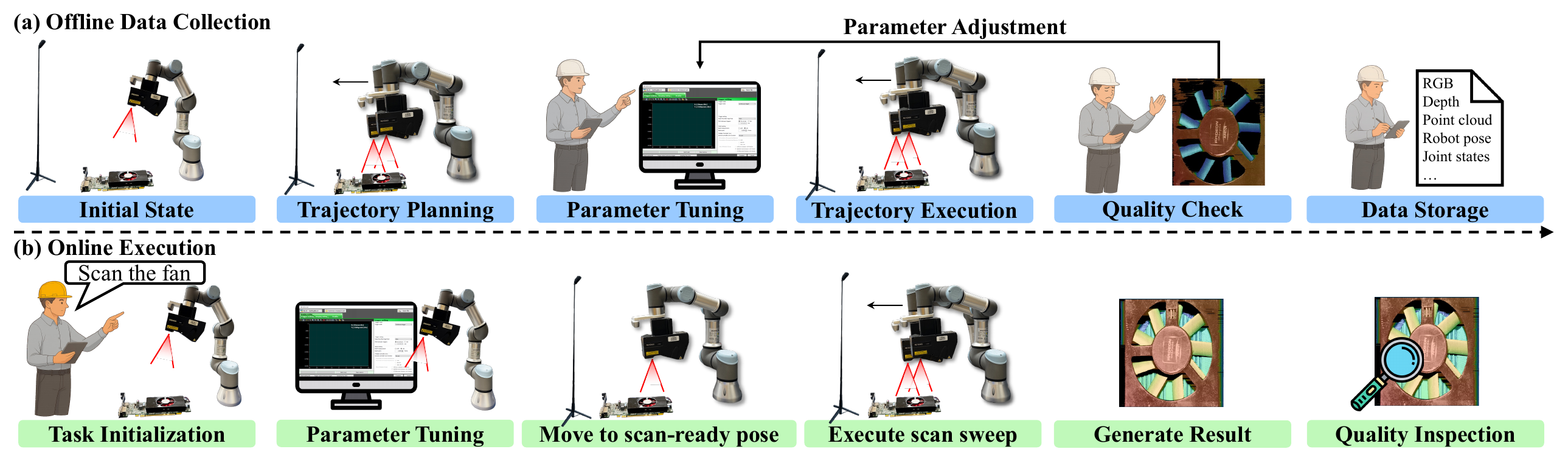}
  \caption{(a) Offline data collection pipeline: workspace setup, trajectory planning/execution with iterative parameter tuning and quality checks, and multimodal data storage. (b) Online execution: given an instruction, the learned policy outputs scan trajectory and sensor parameters; the robot executes the scan sweep and produces surface profiles for quality inspection.}
  \label{data collection}
\end{figure*}

\subsection{Hardware Setup} 

As shown in Fig.~\ref{setup}, the ScanBot system combines a 6-DOF UR3 collaborative robotic arm with a multi-sensor payload for high-precision surface profiling. The end-effector integrates a Keyence LJ-X8200 laser displacement sensor, controlled by an LJ-X8000A controller. An Intel RealSense D435i RGB-D camera is rigidly co-mounted to provide complementary visual and depth data, while a GoPro HERO8 Black records third-person views from a fixed tripod.




\subsection{Data Collection}
\label{sec: data collection}



To reflect real-world industrial inspection scenarios, we design a data collection protocol that supports multi-object, multi-task surface scanning. Each object in the dataset is associated with multiple scanning tasks targeting different geometric features or surface regions, enabling the evaluation of multi-task and instruction-driven learning methods under diverse conditions.

For each trial, we record synchronized multimodal observations (laser profiles, wrist RGB-D, and third-person video) together with robot states and task annotations, and save them as one ScanBot sample.
Although initial scanning paths are pre-defined for each object and feature, our data collection follows a structured protocol illustrated in Fig.~\ref{data collection}(a). We first place the object in the workspace and generate a motion plan for a candidate trajectory aligned with the target feature or region. We then enter an iterative calibration loop that alternates between scan execution and quality checking.
For each trajectory, we refine a subset of laser profiler parameters supported by the LJ-X8000A controller. All parameter values are selected directly from the discrete options provided by the controller firmware, consistent with standard industrial practice, without additional discretization or artificial constraints.
As summarized in Fig.~\ref{task}(a), scanning parameters exhibit different primary drivers. Sampling frequency is mainly instruction-driven as it directly determines spatial resolution, while control light intensity is observation-driven to adapt to surface reflectivity. Measurement range X, CMOS dynamic range, and exposure time depend on both instruction semantics and object appearance, balancing task requirements with sensing stability. For example, coarse instructions (e.g., scanning a large top surface) allow a wider X-range and lower sampling frequency for faster coverage, whereas fine-grained instructions require a narrower range and higher sampling frequency to preserve feature-level detail; reflective materials further require tuning exposure and dynamic range to suppress saturation.

We additionally maintain a nominal standoff close to the scanner reference distance (245\,mm) to keep the profile within the valid measurement range and maximize usable field of view. Finally, we tune robot motion speed and acceleration to match the profiler sampling frequency; insufficient synchronization leads to undersampling or degraded profiles. Once scan quality is verified, we finalize the configuration and record the trial. This process is repeated across all object-task combinations to collect instruction-grounded surface scans for fine-grained evaluation and reasoning.

\subsection{Dataset Composition}


\begin{figure*}[]
  \centering
  \includegraphics[width=\textwidth]{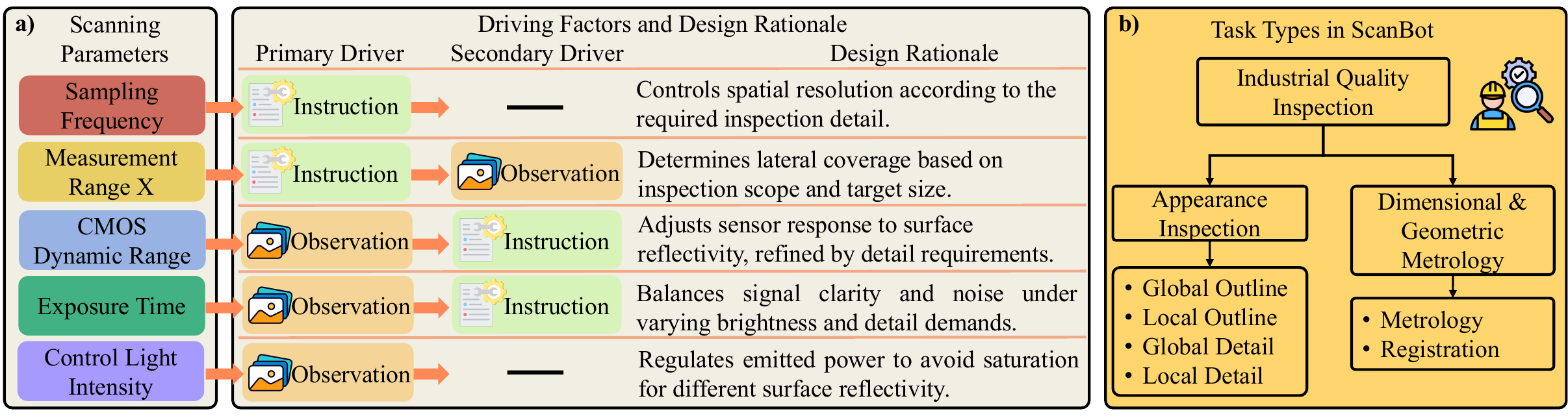} 
  \caption{ScanBot design overview. (a) Five key laser-profiler parameters and their primary drivers (instruction vs. observation) with corresponding design rationale. (b) ScanBot task taxonomy for industrial quality inspection, covering appearance inspection and dimensional/geometric metrology.}
  \label{task}
\end{figure*}

\begin{figure}[]
  \centering
  \includegraphics[width=0.9\linewidth]{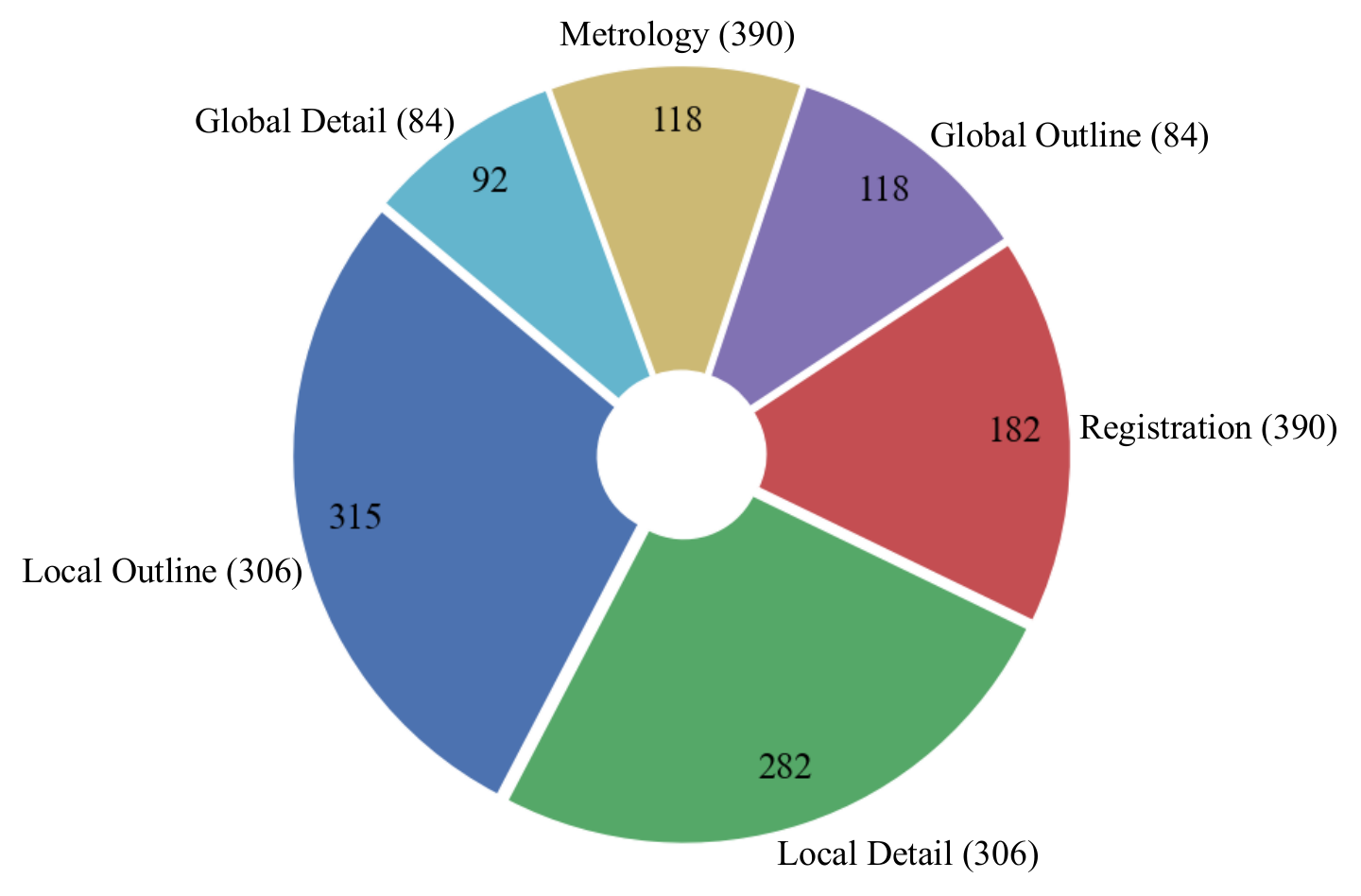}
  \caption{ScanBot task distribution across six categories. Inner-ring numbers denote the number of task instances per category, and outer-ring numbers denote the number of collected scanning paths.}
  \label{data}
\end{figure}

To reflect real industrial quality inspection workflows, ScanBot organizes instruction-conditioned scanning into six task types spanning both appearance inspection and dimensional \& geometric metrology, as summarized in Fig.~\ref{task}(b). Appearance-oriented tasks include Global Outline and Global Detail, which capture object-level shape and salient surface cues, as well as Local Outline and Local Detail, which progressively focus on feature-level geometry and fine-grained surface variations. Metrology-oriented tasks include Metrology, which emphasizes precise geometric measurement on specified regions, and Registration, which targets alignment and consistency across regions, views, or references. Fig.~\ref{data} reports the dataset composition using a two-level ring: the inner-ring numbers indicate the number of task instances in each category, while the outer-ring numbers indicate the number of collected scanning paths in that category. Each task instance is paired with one or more natural-language instructions and executed by one or multiple scanning paths. Every path yields a multimodal sample with synchronized sensor streams and detailed metadata, including the instruction text, task type/ID, robot joint states, end-effector poses, and scanner settings.

\begin{figure*}[]
  \centering
  \includegraphics[width=\textwidth]{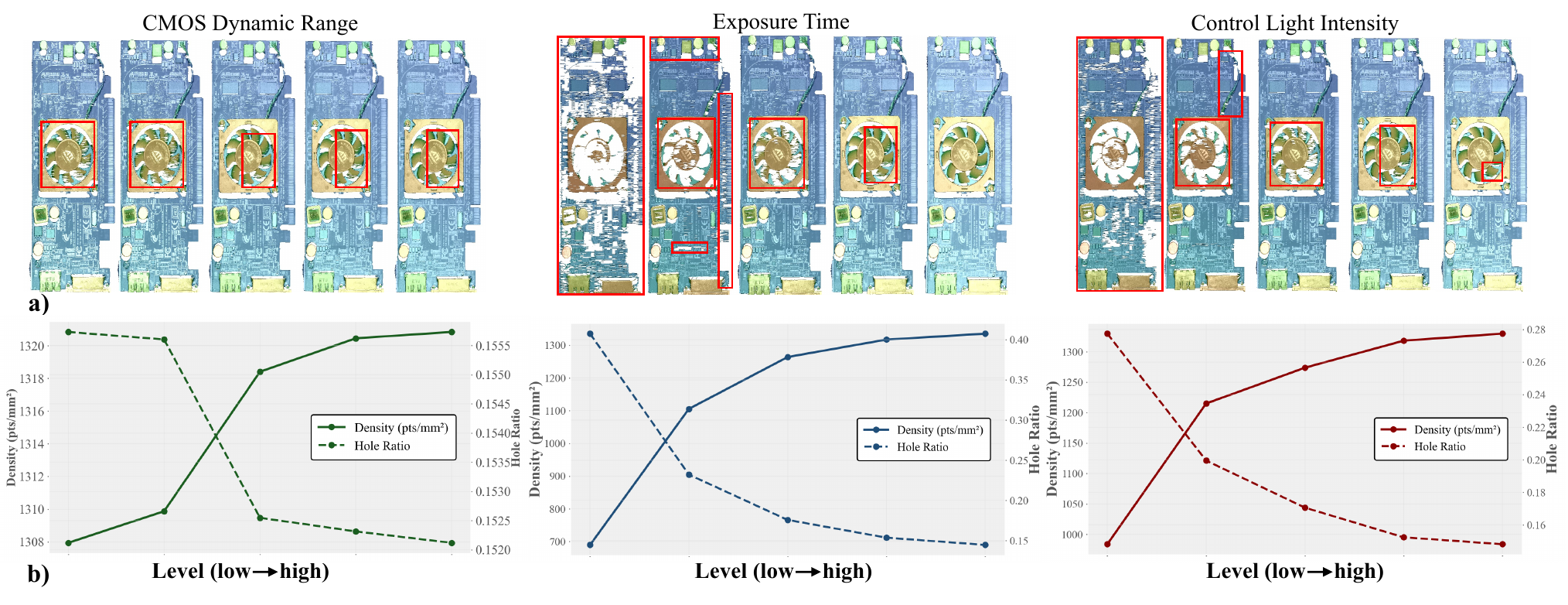} 
  \caption{Effect of scanner parameter variation on scanning results.
(a) Representative point-cloud projections under increasing parameter levels, the regions highlighted by red boxes indicate representative areas where point dropout and hole artifacts are most visually pronounced for qualitative comparison.
(b) Corresponding quantitative trends in sampling density and hole ratio, showing improved coverage and reduced dropout with proper parameter tuning.}
  \label{exp1}
\end{figure*}

\section{BASELINES}
\label{sec:baselines}

To benchmark the proposed ScanBot dataset and validate its utility for learning high-precision scanning skills, we establish a two-stage evaluation protocol: (1) \textit{Parameter Prediction} and (2) \textit{Scan Motion Generation}. In this section, we describe the baseline methods and our proposed learning-based reference models for each stage.

\subsection{Stage I: Parameter Inference Models}

We evaluate scanner parameter inference using a comprehensive set of offline baselines designed to isolate the contribution of instruction semantics, visual observations, and their multimodal integration, as well as to assess the capabilities of state-of-the-art multimodal large language models. 

\textbf{Rule-based baseline:} Hand-crafted heuristics for scanner parameter selection, mimicking industrial tuning from inspection intent and coarse appearance cues.

\textbf{Instruction-only baseline:} We predict parameters from language alone (no vision) using KNN on CLIP text embeddings, measuring how much intent alone explains parameter choices.

\textbf{Observation-only baseline:} We predict from RGB alone (no instruction) using ViT, isolating the effect of appearance cues without task semantics.

\textbf{Fusion baseline:} We perform feature-level fusion by concatenating RGB and instruction embeddings, followed by a three-layer MLP head to jointly infer scanning parameters.

\textbf{Multimodal large language models:}  Zero-shot parameter inference from image--instruction pairs using Qwen3-VL variants (Instruct/Thinking).

\subsection{Stage II: Scan Motion Generation}
Conditioned on the observation and instruction, the robot must generate a smooth scanning trajectory $\tau$. We use the following approaches as diagnostic references to expose sensing-critical bottlenecks in scanning:

\textbf{Industrial Reference:} 
A conventional industrial setup, referred to as Fixed-Template Scanning (FTS), in which objects are placed within a predefined workcell region and scanned using a fixed, pre-designed trajectory.

\textbf{Single-Task Imitation Learning Baselines:} 
We train task-specific policies using demonstration data from ScanBot. In particular, we evaluate Action Chunking Transformer (ACT) \cite{zhao2023learning} and Diffusion Policy (DP) \cite{chi2025diffusion} as representative imitation learning methods. 

\textbf{VLA Models:} 
We evaluate state-of-the-art VLA models, including $\pi_0$ \cite{black2026pi0visionlanguageactionflowmodel}and $\pi_{0.5}$ \cite{intelligence2025pi05visionlanguageactionmodelopenworld}.

\section{Experiments}
\label{experiments}


The goal of our experiments is to validate the utility of the ScanBot benchmark for learning high-precision, instruction-conditioned surface scanning skills. As illustrated in Fig.~\ref{data collection} (b), our evaluation covers the entire perception $\rightarrow$ planning $\rightarrow$ execution loop: the model first observes multi-modal sensor data from an unseen object, then interprets an instruction, selects suitable scanner parameters, plans a trajectory, and finally executes a scan to produce high-resolution surface profiles. We aim to demonstrate that, unlike standard manipulation benchmarks, ScanBot provides the necessary multi-modal supervision to master sensing-driven behaviors, where success depends on both precise inference of scanner parameters and the geometric stability of the executed motion.

Our experiments are designed to answer the following questions:
\begin{enumerate}[]
    \item Can we infer task-appropriate scanner parameters from object appearance and task instructions?
    \item Can we generate stable and feasible scan motions that adequately cover the target well without extra moves?
\end{enumerate}

\begin{table*}[h!]
\centering
\caption{Stage-I parameter inference results on ScanBot. We report exact accuracy and macro F1 for five scanner parameters.}
\label{tab:question1}
\resizebox{\textwidth}{!}{
\begin{tabular}{l|cc|cc|cc|cc|cc|cc}
\toprule
& \multicolumn{2}{c|}{\textbf{Sampling Frequency}} 
& \multicolumn{2}{c|}{\textbf{Measurement Range X}} 
& \multicolumn{2}{c|}{\textbf{Exposure Time}} 
& \multicolumn{2}{c|}{\textbf{CMOS Dynamic Range}} 
& \multicolumn{2}{c|}{\textbf{Control Light Intensity}} 
& \multicolumn{2}{c}{\textbf{Average}} \\
\textbf{Method} 
& Exact Acc& F1
& Exact Acc& F1
& Exact Acc& F1
& Exact Acc& F1
& Exact Acc& F1
& Exact Acc& F1 \\ 
\midrule

\multicolumn{13}{l}{\textbf{A. Rule-based Baseline}} \\
Rule-based Heuristic 
& 41.7$\pm$2.0 & 22.5$\pm$1.8 
& 53.5$\pm$4.8 & 23.2$\pm$1.4 
& 43.0$\pm$2.5 & 20.6$\pm$1.2 
& 54.7$\pm$4.5 & 24.7$\pm$1.4 
& 49.7$\pm$2.5 & 22.6$\pm$1.0 
& 48.5$\pm$3.3 & 22.7$\pm$1.4 \\
\midrule

\multicolumn{13}{l}{\textbf{B. Instruction-Only Baseline}} \\

KNN
& \textbf{91.7$\pm$2.0} & \textbf{91.4$\pm$1.8} 
& 82.2$\pm$2.6 & 79.4$\pm$4.2 
& 73.5$\pm$4.7 & 72.7$\pm$5.0 
& 81.7$\pm$2.8 & 77.8$\pm$3.3 
& 80.0$\pm$3.1 & 78.0$\pm$2.7 
& 81.8$\pm$3.0 & 79.9$\pm$3.4 \\
\midrule

\multicolumn{13}{l}{\textbf{C. Observation-Only Baseline}} \\

ViT
& 41.8$\pm$4.7 & 31.6$\pm$1.3 
& 75.9$\pm$2.0 & 76.6$\pm$2.4 
& 71.4$\pm$4.5 & 71.9$\pm$3.1 
& 75.0$\pm$2.7 & 76.7$\pm$2.7 
& \textbf{100.0$\pm$0.0} & \textbf{100.0$\pm$0.0} 
& 72.8$\pm$2.8 & 71.4$\pm$1.9 \\
\midrule

\multicolumn{13}{l}{\textbf{D. Fusion Baseline}} \\
DNN
& 82.3$\pm$3.2 & 81.3$\pm$4.2 
& \textbf{94.1$\pm$2.1} & \textbf{94.3$\pm$1.5} 
& \textbf{83.5$\pm$3.0} & \textbf{83.7$\pm$2.7} 
& \textbf{84.0$\pm$4.2} & \textbf{82.7$\pm$4.7} 
& 99.2$\pm$0.7 & 99.0$\pm$0.9 
& \textbf{88.6$\pm$2.6} & \textbf{88.2$\pm$2.8}\\
\midrule

\multicolumn{13}{l}{\textbf{E. Multimodal Large Language Models}} \\
Qwen3-VL-4B-Instruct
& 58.7$\pm$3.3 & 53.8$\pm$3.6 
& 58.3$\pm$3.4 & 46.4$\pm$3.1 
& 25.1$\pm$1.6 & 19.6$\pm$1.6 
& 58.3$\pm$2.3 & 46.1$\pm$1.9 
& 32.9$\pm$3.2 & 30.1$\pm$3.4 
& 46.7$\pm$2.8 & 39.2$\pm$2.7 \\

Qwen3-VL-4B-Thinking
& 63.2$\pm$3.0 & 61.6$\pm$3.0 
& 61.9$\pm$2.5 & 51.1$\pm$1.8 
& 35.0$\pm$1.8 & 31.3$\pm$1.8 
& 54.2$\pm$2.2 & 44.8$\pm$2.7 
& 26.5$\pm$2.3 & 26.6$\pm$2.3 
& 48.2$\pm$2.4 & 43.1$\pm$2.3 \\

Qwen3-VL-8B-Instruct
& 65.4$\pm$2.1 & 49.1$\pm$2.0 
& 63.3$\pm$2.3 & 45.8$\pm$1.3 
& 35.4$\pm$2.3 & 29.7$\pm$2.8 
& 51.2$\pm$2.2 & 40.3$\pm$2.5 
& 33.8$\pm$2.3 & 28.5$\pm$1.7 
& 49.8$\pm$2.2 & 38.7$\pm$2.1  \\

Qwen3-VL-8B-Thinking
& 63.5$\pm$2.9 & 58.3$\pm$4.1 
& 63.2$\pm$2.2 & 47.2$\pm$1.6 
& 33.2$\pm$3.9 & 33.2$\pm$3.8 
& 45.0$\pm$3.3 & 38.2$\pm$3.1 
& 44.6$\pm$1.5 & 34.4$\pm$1.7 
& 49.9$\pm$2.8 & 42.3$\pm$2.9\\

\bottomrule
\end{tabular}
}
\end{table*}

\subsection{Can we infer task-appropriate scanner parameters from object appearance and task instructions?}

\noindent\textbf{Setup.}
We study scanner parameter inference from an image--instruction pair. Each sample consists of an RGB observation, a natural-language instruction, and a vector of parameter levels. ScanBot provides 1107 instruction--observation--parameter triplets, which we split into 80\%/20\% train/test at the trajectory level to avoid overlap between splits. We evaluate all the models under the same protocol. Performance is measured using exact accuracy and macro-F1. For stochastic learning-based baselines, we run each model 10 times with different random seeds and report mean performance (with variance across runs).

\noindent\textbf{Results.}
Before benchmarking inference models, we validate that ScanBot parameter levels induce systematic changes in reconstruction quality. We sweep three signal-level settings—CMOS dynamic range, exposure time, and control light intensity—by varying one parameter from low to high while keeping the scan motion and remaining settings fixed. Higher levels yield more complete point-cloud projections with fewer structured dropouts (Fig.~\ref{exp1}a), and quantitatively increase sampling density while reducing hole ratio across all sweeps (Fig.~\ref{exp1}b), confirming that the labels correspond to physically meaningful sensing regimes that affect geometric coverage and signal stability.

We then assess whether task-appropriate parameters can be inferred from instruction semantics, visual evidence, and their fusion (Table~\ref{tab:question1}). 
Instruction-only prediction remains strong on intent-dominated settings: the KNN baseline achieves high accuracy on sampling frequency and measurement range X, indicating that inspection intent provides a reliable prior for coarse sensing regimes. Observation-only models are weaker on parameters that depend on task semantics, but they excel on visually governed settings: ViT achieves near-perfect performance on control light intensity, consistent with its dependence on surface reflectivity cues. Multimodal fusion is the most consistent across parameters; the fusion DNN achieves the best overall performance and notably improves coupled settings that require both intent and appearance, especially exposure time, while maintaining strong results on the remaining parameters. In contrast, the rule-based heuristic underperforms across most parameters, suggesting that hand-crafted rules fail to capture the structured coupling present in ScanBot. Finally, zero-shot multimodal large language models (Qwen3-VL variants) lag behind the strongest learned baselines, with the largest gap on exposure time, highlighting that discrete, constraint-aware parameter selection benefits from task-specific learning beyond prompting.

\subsection{Can we generate stable and feasible scan motions that adequately cover the target well without extra moves?}

\begin{table}[t]
\centering
\caption{Task-level sensing viability for laser profiler inspection. 
Higher is better ($\uparrow$) for Target Coverage (TC) and Valid Data Rate (VDR); lower is better ($\downarrow$) for Root Mean Square Scan Deviation (RMS SD) and Max Gap (MG).}
\label{tab:sensing_viability}
\setlength{\tabcolsep}{3pt} 
\scalebox{\tblscale}{
\begin{tabular}{l|c|c|c|c}
\toprule
\textbf{Method} 
& \makecell{\textbf{TC} \\ \textbf{(\%) $\uparrow$}}
& \makecell{\textbf{RMS SD} \\ \textbf{(mm) $\downarrow$}} 
& \makecell{\textbf{VDR} \\ \textbf{(\%) $\uparrow$}}
& \makecell{\textbf{MG} \\ \textbf{(mm) $\downarrow$}} \\
\midrule
\multicolumn{5}{l}{\textbf{A. Industrial Reference}} \\
FTS 
& 99.92$\pm$0.01 & 0.05$\pm$0.01 & 100$\pm$0.00 & 0.02$\pm$0.00 \\
\midrule
\multicolumn{5}{l}{\textbf{B. Single Task Imitation Learning Methods}} \\
ACT 
& 39.48$\pm$2.75 & 62.97$\pm$0.16 & 93.27$\pm$7.43 & 7.44$\pm$1.87 \\
DP
& 40.20$\pm$2.16 & 62.43$\pm$0.47 & 89.33$\pm$0.50 & 7.50$\pm$0.71 \\
\midrule
\multicolumn{5}{l}{\textbf{C. Vision-Language-Action Models}} \\
$\pi_0$ 
& 19.66$\pm$0.20 & 60.78$\pm$0.02 & 72.81$\pm$0.02 & 64.72$\pm$0.04 \\
$\pi_{0.5}$
& 24.64$\pm$0.14 & 37.30$\pm$0.04 & 69.51$\pm$0.02 & 45.98$\pm$0.29 \\
\bottomrule
\end{tabular}
}
\end{table}

\begin{table}[t]
\centering
\caption{Kinematic stability diagnostics for laser profiler inspection. 
Lower is better ($\downarrow$) for all metrics, including Mean Absolute Jerk (MAJ), Stand-off Distance Deviation (SDD), and Z-axis Angular Jitter (Z-AJ).}
\label{tab:kinematic_stability}
\setlength{\tabcolsep}{3pt} 
\scalebox{\tblscale}{
\begin{tabular}{l|c|c|c}
\toprule
\textbf{Method} 
& \makecell{\textbf{MAJ} \\ \textbf{(rad/s$^3$) $\downarrow$}}
& \makecell{\textbf{SDD} \\ \textbf{(mm) $\downarrow$}}
& \makecell{\textbf{Z-AJ} \\ \textbf{(deg/s) $\downarrow$}} \\
\midrule
\multicolumn{4}{l}{\textbf{A. Industrial Reference}} \\
FTS 
& 0.42$\pm$0.00 & 0.02$\pm$0.00 & 0.03$\pm$0.00 \\
\midrule
\multicolumn{4}{l}{\textbf{B. Single Task Imitation Learning Methods}} \\
ACT 
& 0.52$\pm$0.03 & 9.92$\pm$0.93 & 9.32$\pm$0.32 \\
DP
& 12.26$\pm$0.20 & 9.78$\pm$0.04 & 11.49$\pm$0.33 \\
\midrule
\multicolumn{4}{l}{\textbf{C. Vision-Language-Action Models}} \\
$\pi_0$ 
& 7.72$\pm$0.01 & 26.83$\pm$0.01 & 4.92$\pm$0.01 \\
$\pi_{0.5}$
& 5.69$\pm$0.01 & 22.86$\pm$0.04 & 4.96$\pm$0.01 \\
\bottomrule
\end{tabular}
}
\end{table}

\begin{figure}[]
  \centering
  \includegraphics[width=0.5\textwidth]{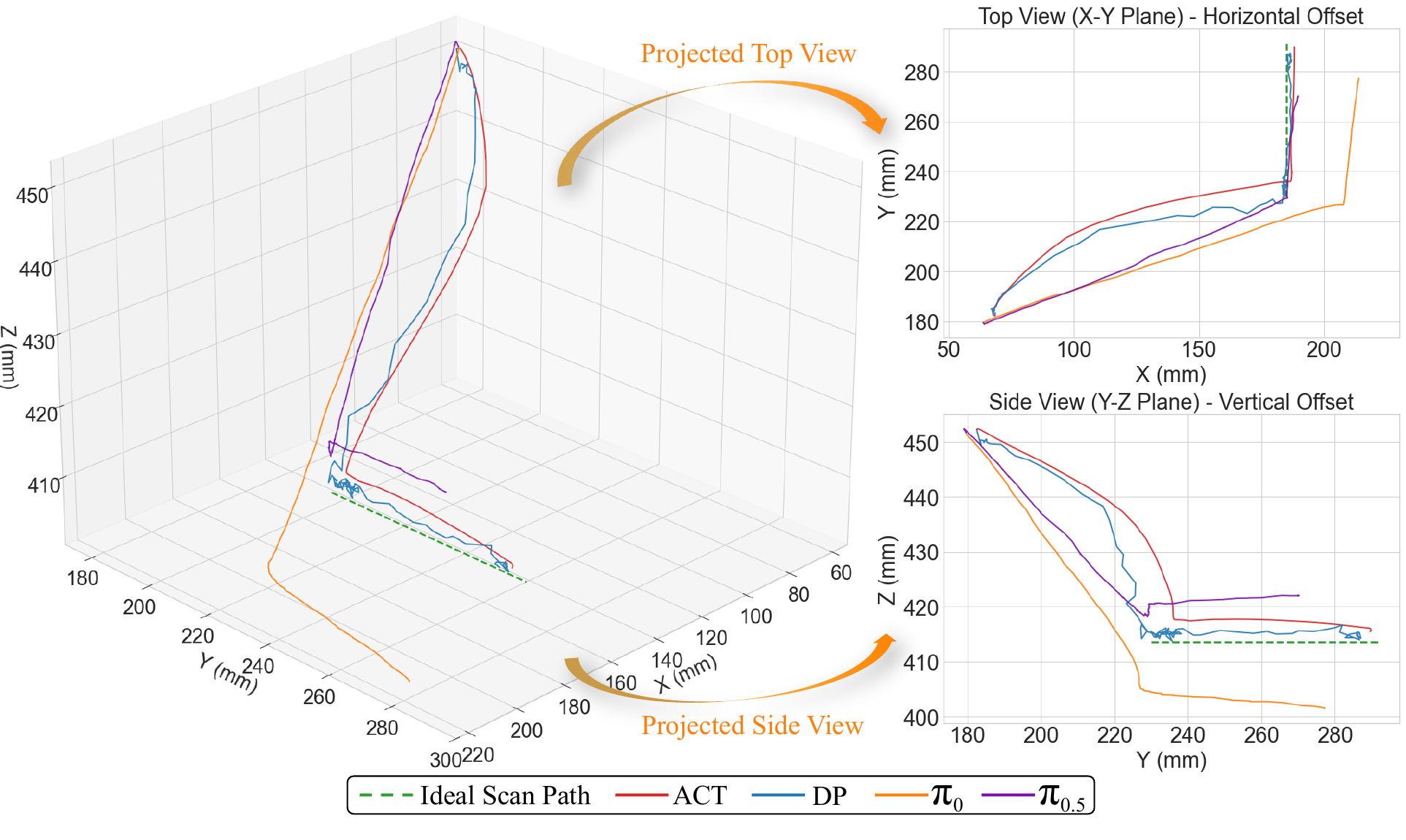} 
  \caption{Representative scan trajectories for one fixed task instance (same object/feature under an identical sweep specification). Left: 3D end-effector trajectories. Right: projections onto the XY (lateral alignment) and YZ (stand-off) planes in a task-aligned frame where the ideal scan line defines the nominal direction.}
  \label{exp2}
\end{figure}

\noindent\textbf{Setup.}
To ensure a fair, data-matched comparison, we train policies \emph{per task type} using only demonstrations collected for that task: ACT and Diffusion Policy are trained from scratch, while $\pi_0$ and $\pi_{0.5}$ are fine-tuned via LoRA on the same task-specific data. All training and inference are performed on a single NVIDIA RTX PRO 6000 Blackwell GPU.

Precision scanning differs fundamentally from manipulation because success is not defined by a discrete terminal state. A scan is only useful if the robot continuously tracks a time-parameterized trajectory aligned with the instructed target region under strict geometric and kinematic constraints. We therefore evaluate all methods on the same set of tasks and run experiments separately for each task type under identical target definitions, scan extents, and sensing constraints. Rather than a binary success signal, we adopt two complementary evaluation levels: Table~\ref{tab:sensing_viability} reports task-level sensing viability, and Table~\ref{tab:kinematic_stability} provides kinematic stability diagnostics. Task-level metrics include \emph{Target Coverage} (TC), the fraction of instructed scan extent covered by valid samples (computed by binning the ideal extent along the scan direction and counting visited bins); \emph{Root Mean Square Scan Deviation} (RMS SD), the RMS lateral (XY) distance between executed samples and the ideal scan line; \emph{Valid Data Ratio} (VDR), the fraction of timesteps whose stand-off error along tool $Z$ stays within $z_{\mathrm{limit}}$; and \emph{Max Gap} (MG, mm), the largest uncovered interval along the scan extent when considering only valid samples. Kinematic metrics include \emph{Mean Absolute Jerk} (MAJ), the mean magnitude of joint-position jerk; \emph{Stand-off Distance Deviation} (SDD), the mean absolute deviation from the reference stand-off along tool $Z$; and \emph{Z-axis Angular Jitter} (Z-AJ), the temporal variation of angular velocity about the tool's local $Z$ axis.

\noindent\textbf{Results.}
Table~\ref{tab:sensing_viability} shows that the industrial reference (FTS) achieves near-ideal sensing viability, with high TC, small RMS SD, a high VDR, and a minimal MG, thereby anchoring an upper bound on stability under our profiler constraints. However, FTS is a fixed, hand-engineered routine and is not instruction-adaptive; it prioritizes conservative, uniform sweeps that may extend beyond the instruction-implied extent and thus does not target feature-specific regions.
In contrast, learned baselines exhibit sharply reduced TC and substantially larger MG even when their trajectories appear to traverse the scan extent, indicating that failures are driven by intermittent loss of usable samples rather than simply stopping early. This behavior is consistent with real laser profiling, where small pose jitter and stand-off fluctuations can push measurements outside the effective range, producing dropouts that translate into missing points and large uncovered intervals. The imitation learning baselines also incur noticeably larger RMS SD, reflecting lateral drift from the intended scan line that further reduces effective coverage and introduces redundant off-target motion. Their VDR remains relatively high, suggesting that many samples are still returned, but a nontrivial fraction is collected away from the target extent and contributes less to inspection completeness. The VLA baselines degrade further, showing the lowest coverage and largest gaps, consistent with stronger instability and weaker geometric grounding that amplify dropouts into contiguous missing regions.

Table~\ref{tab:kinematic_stability} attributes these coverage failures to long-horizon constraint-tracking errors. The industrial template baseline maintains low MAJ, SDD, and Z AJ, matching its strong viability in Table~\ref{tab:sensing_viability}. Among learned methods, ACT tends to produce smoother joint trajectories yet exhibits elevated SDD and Z AJ, indicating insufficient regulation of stand-off distance and tool orientation that is critical for preventing dropouts. Diffusion Policy shows the opposite pattern, with larger MAJ together with high SDD, suggesting that high-frequency motion irregularities and distance violations occur simultaneously; this combination is particularly harmful for laser profiling and explains the large gaps and reduced coverage. The VLA baselines show the most pronounced distance violations, indicating difficulty maintaining the stand-off constraint over extended trajectories, which aligns with their low sensing viability and large MG.

To better visualize these failure modes, Fig.~\ref{exp2} plots representative trajectories from the learned policies for the same task instance (same object/feature) under an identical sweep specification (i.e., the same reference scan line and nominal extent). For a fair comparison, we express all trajectories in a task-aligned frame where the ideal scan line defines the nominal scan direction. Under this normalization, policies share a similar coarse shape since feasible motions lie on a narrow manifold constrained by the scan line, stand-off, collision avoidance, and actuator limits. The differences thus reflect tracking quality: the XY and YZ projections reveal lateral drift, stand-off deviations, and off-path excursions. These geometric errors align with the reduced TC and increased MG in Table~\ref{tab:sensing_viability}, supporting the diagnosis in Table~\ref{tab:kinematic_stability} that imperfect distance tracking and motion instability cause intermittent dropouts and contiguous missing regions.

Overall, these results indicate that the main bottleneck of learned policies is not the nominal scan geometry, but long-horizon sensor-viable constraint tracking. In real laser scanning, small stand-off and orientation errors readily cause dropouts, which accumulate into large gaps and reduced coverage. This motivates VLA-style scanning policies that explicitly represent and regulate sensing-critical constraints (e.g., stand-off and incidence angle) throughout execution.

\section{Conclusion}
\label{conclusion}

In this paper, we established ScanBot as a step toward instruction-conditioned, tool-mediated sensing, where performance is governed not only by task completion but also by the physical constraints of industrial laser profiling. By framing scanning as a two-stage problem that separates sensor configuration from motion generation, we provide a practical way to diagnose where current systems fail when asked to follow fine-grained scan instructions under sub-millimeter precision requirements. Our benchmarks indicate that, even with recent advances in multimodal and VLA models, generating stable, feasible, and coverage-sufficient scan motions remains a major bottleneck, pointing to the need for robot models that couple perception, language grounding, and continuous control more tightly. ScanBot also has limitations. 
The current setting emphasizes flat-surface trajectories, adopts a predominantly single-pass, feedforward execution protocol without online correction, and focuses on fixed sensing parameters during each scan. Future work will expand ScanBot to curved and irregular geometries, incorporate online pose and parameter refinement for adaptive execution, and introduce multi-pass protocols that better reflect real inspection practice.


\bibliographystyle{IEEEtran}
\bibliography{refs}


\addtolength{\textheight}{-12cm}   

\end{document}